\documentclass[10pt,twocolumn,letterpaper]{article}

\usepackage{cvpr}              

\usepackage[dvipsnames]{xcolor}

\usepackage[pagebackref,breaklinks,colorlinks]{hyperref}

\usepackage{graphicx, caption}
\usepackage{tabularx}
\usepackage{booktabs}
\usepackage{multirow}
\usepackage{xspace}
\usepackage{float} 
\DeclareMathOperator*{\argmin}{arg\,min}
\def\paperID{325} 
\def\confName{3DV\xspace}
\def\confYear{2024\xspace}

\title{PRAGO: Differentiable Multi-View Pose Optimization From Objectness Detections\thanks{This project has received funding from the European Union's Horizon research and innovation programme under grant agreement No 870743 and under grant agreement No 101079995.}}

\author{Matteo Taiana, Matteo Toso, Stuart James and Alessio {Del Bue}\\
Pattern Analysis and Computer Vision (PAVIS), Istituto Italiano di Tecnologia (IIT)\\
Genoa, Italy\\
{\tt\small \{name.surname\}@iit.it}
}

\begin{document}
\maketitle
\newcommand{\NETNAME}{PoserNet++}
\newcommand{\NETNAMEwithSpace}{PoserNet++ }
\newcommand{\APPROACHNAME}{PRAGO}
\newcommand{\APPROACHNAMEwithSpace}{PRAGO }
\begin{abstract}
Robustly estimating camera poses from a set of images is a fundamental task which remains challenging for differentiable methods, especially in the case of small and sparse camera pose graphs.  
To overcome this challenge, we propose Pose-refined Rotation Averaging Graph Optimization (PRAGO). From a set of objectness detections on unordered images, our method reconstructs the rotational pose, and in turn, the absolute pose, in a differentiable manner benefiting from the optimization of a sequence of geometrical tasks.
We show how our objectness pose-refinement module in PRAGO is able to refine the inherent ambiguities in pair-wise relative pose estimation without removing edges and avoiding making early decisions on the viability of graph edges. PRAGO then refines the absolute rotations through iterative graph construction, reweighting the graph edges to compute the final rotational pose, which can be converted into absolute poses using translation averaging. 
We show that PRAGO is able to outperform non-differentiable solvers on small and sparse scenes extracted from 7-Scenes achieving a relative improvement of $21\%$ for rotations while achieving similar translation estimates.

\end{abstract}    
\section{Introduction}
\label{sec:intro}

Most current Structure from Motion (SfM) approaches \cite{schonberger2016structure,farenzena2009structure,moulon2016openmvg,snavelySIGGRAPH06photo} require an initial robust  estimation of absolute camera poses from multi-view images. This first step is critical, as it is the basis for obtaining a coarse 3D point cloud by triangulation and then feeding standard non-linear solvers \cite{triggs2000bundle,agarwal2012ceres} that estimate a refined set of camera poses (rotation and translation) and 3D geometry. Given the nonlinearity of the problem, an initialization farther away from the optimal solution will increase the chances of obtaining poor-quality reconstructions or failing entirely.

As SfM pipelines evolved through time, a new trend is to inject semantics in multi-view problems. Semantic and object SfM approaches~\cite{bao2011semantic,crocco2016structure} provide not only a raw 3D point cloud, but they are able to localize objects in 3D and to estimate their pose. Objects semantic is also used in SLAM methods \cite{galvez2016real,bowman2017probabilistic,mccormac2018fusion++,nicholson2018quadricslam,yang2019cubeslam} to better localize the camera and objects in structured scenes and to build 3D scene graphs for improved scene understanding or robotic navigation \cite{rosinol2021kimera}.   

On a similar trend, this paper explores how the semantic information of objects can help to estimate better absolute camera rotations and translations in an SfM scenario. In particular, we address one of the hardest cases of SfM, small-scale scenes with fewer cameras and wider baselines, as most approaches make assumptions of large and dense camera pose graphs, with relatively short baselines or expendable images. Whenever few cameras are available, obtaining a robust initialization of camera poses is harder, as the 2D correspondences are fewer and noisier, so even outliers rejection methods \cite{wilson_eccv2014_1dsfm} struggle to find good inliers or eliminate too many cameras/points from the initial set.

Previous work has shown that objectness-based image detections can refine relative poses~\cite{taiana2022posernet} by reasoning on higher-level semantic information. The method defines a pose graph where each node represents a camera and its corresponding detections, while edges relate together nodes that have matching detections in multi-view. Taiana~\etal~\cite{taiana2022posernet} proposed a novel graph-based computational approach embedding the images' detections on the nodes of the graph, and the aggregate of predictions of the refined relative pose on the edges. Message passing propagated information in an MLP-style alternating updates between nodes and edges. 
This approach successfully improved relative poses, but was trained independently from the rest of the pipeline, and relied on an optimization algorithm to perform motion averaging (\ie EIG-SE3~\cite{arrigoni2016spectral}).

In contrast, we propose PRAGO, a differentiable pose estimation method,  which focuses on small and sparse pose graphs where camera outlier rejection (e.g. 1DSfM \cite{wilson_eccv2014_1dsfm}) might be unpractical. 
PRAGO uses noisy pairwise camera poses estimated from an objectness-based approach, refines the relative rotations and then integrates rotation averaging to estimate absolute camera poses. As we show experimentally, optimizing these tasks together can provide a significant improvement. As PRAGO does not apply camera outlier rejection but instead focuses on the consistency of the matched objectness detections, we avoid making decisions early on the viability of the pose graph, which makes the method highly suitable for small and sparse graphs where the removal of even a couple of cameras would have a detrimental effect in later steps. 
Specifically, we propose a graph where edges encode the initial relative pose and bounding box on each respective image, while the node simply store the intrinsic parameters for each camera. Through message-passing updates between edges and nodes in the graph, the nodes are updated and a new pose is regressed as input to motion averaging. At this point, PRAGO iteratively refines the pose graph, re-weighting the contribution of rotational information mimicking an optimization process, where the RAGO \cite{li2022rago} blocks incrementally improve the accuracy of the estimates. The rotations are combined with the refined translations to recover absolute camera poses.
To understand the importance of pose refinement within PRAGO, we provide an analysis of the orientation error and identify that it frequently fixes chirality errors. In addition, we compare to state-of-the-art combinations of independent methods, both differentiable or optimization-based.  

We make three major contributions: 
\emph{i)} we introduce the PRAGO network for differentiable pose estimation with objectness detections where the back-propagation improves performance in contrast to non-differential pipelines;
\emph{ii)} we revise the objectness pose refinement ~\cite{taiana2022posernet} which leads to improved accuracy;
and \emph{iii)} we provide an analysis of how objectness-based pose refinement improves relative poses and its positive effect on chirality. 

\section{Related work}
\label{sec:formatting}

In this section, we provide an introduction to relative pose estimation (Sec.~\ref{sec:litrev_relativepose}) and motion averaging (Sec.~\ref{sec:motionavglit}), a more detailed discussion on methods dedicated to averaging the rotation (Sec.~\ref{sec:litrev_rotationavg}) and translation (Sec.~\ref{sec:litrev_translationavg}) components of the pose. For a full review of SfM see Bianco~\etal~\cite{bianco}. 

\subsection{Relative Pose Estimation}\label{sec:litrev_relativepose}
The first step for obtaining a set of absolute camera poses is to estimate pairwise relative poses from 2D image correspondences.
Given a pair of images, their relative pose can be easily estimated via the five-point algorithm~\cite{1288525, ICPR2006579} or the eight-point algorithm~\cite{601246}. Outlier-rejection strategies such as RANSAC~\cite{fischler1981random} are then used to obtain a robust estimate with a desirable low noise. 
Computing a rotation matrix and a translation unit vector from the Essential matrix presents an inherent ambiguity, though, as two possible solutions exist~\cite{1288525}. One of the two solutions per view is chosen based on chirality considerations (triangulated 3D points) \cite{hartley2003multiple}. Picking the incorrect combinations results in outlier errors that can be discarded in the following steps.

More recent works on relative pose estimation leverage deep learning to directly predict a relative pose given two images~\cite{LI2021134,barroso2023two}, or a set of feature matches and a sparse scene reconstruction~\cite{moran2021deep}. Learning-based models can also exploit cues like light source directions, vanishing points, and scene symmetries~\cite{Cai2021Extreme}, or use Neural Radiance Fields (NeRF)~\cite{mildenhall2020nerf} to predict relative poses~\cite{yen2020inerf}. 
Finally, works like PoserNet~\cite{taiana2022posernet} combine the two approaches, initializing the relative pose via geometrical methods, that are more efficient and do not require training, and then training a model to predict and correct the noise on the predictions.

Noisy relative poses, and especially outliers, have a big impact on motion averaging algorithms, so such algorithms are often designed to explicitly deal with these kinds of errors. Classical approaches usually address this by identifying the outliers measurements by imposing cycle consistency, \ie chaining relative transformations along a closed loop~\cite{5539801}, 
or by projecting the scene graph in a lower dimension and solving a minimum feedback arc set problem to identify the outliers, like 1DSfM~\cite{wilson_eccv2014_1dsfm}. Considering that in the case of sparse or small pose graphs rejecting relative poses can lead to catastrophic results (a graph with multiple components does not contain enough information for estimating coherent absolute poses), we design a pose refinement module to learn to refine, rather than to reject them.

\begin{figure*}[t]
    \centering
    \includegraphics[width=\linewidth]{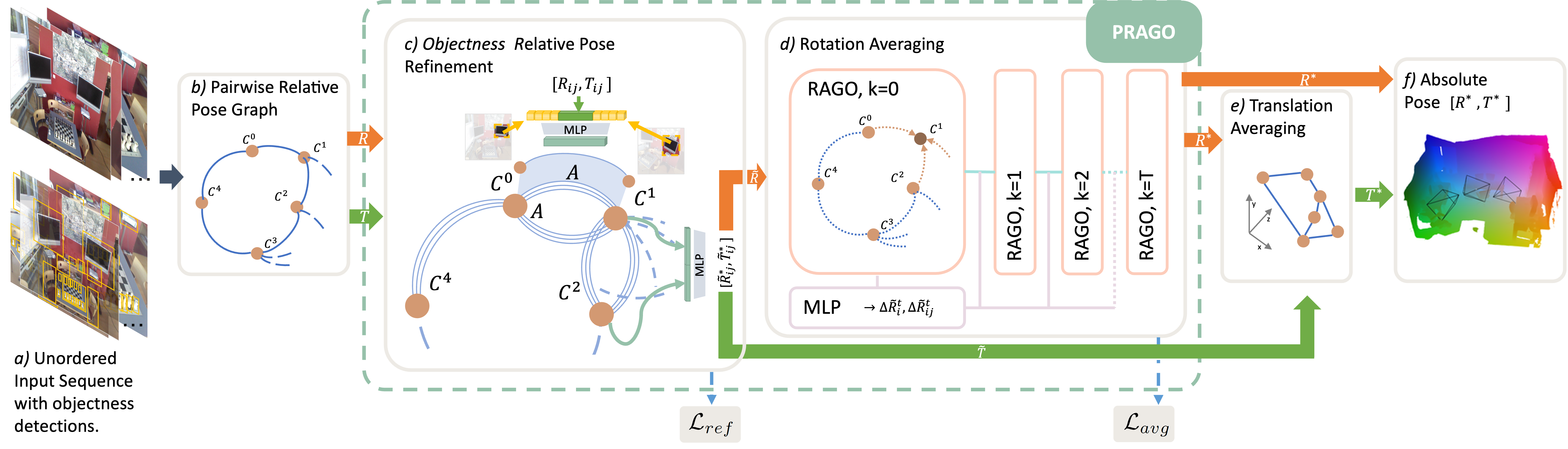}
    \caption{From a set of unordered images \emph{a)}, we compute objectness detections and construct an initial relative pose graph from \emph{b)} the 5-Point pose estimates. \emph{PRAGO} refines the relative pose using the objectness detections resolving, among others, the chirality issues before inferring the rotational poses using RAGO blocks which iteratively infer the rotational pose by reweighting the graph edges. The absolute rotations and refined translations are then combined in \emph{e)} translation averaging using BATA to construct the \emph{f)} absolute camera poses.}
    
    
    \label{fig:main_fig}
\end{figure*}

\subsection{Motion Averaging}\label{sec:motionavglit}
Relative poses are combined to extract absolute poses through optimization algorithms that simultaneously optimize over the orientation and translation components of the pose~\cite{moulon2016openmvg, arrigoni2016spectral, Arrigoni_2021_ICCV, 9577752, yang2021end, yew2021learning}. 
Other optimization-based approaches instead decouple rotation estimation from translation and scene reconstruction, and focus on either rotation~\cite{Lee_2021_CVPR,purkait2020neurora} or translation~\cite{zhuang2018baseline} averaging.
Finally, there are methods that focus on object detections, performing diffusion and clustering to compute
the candidate poses of the input objects, optimizing the best pose for each object and then refining the camera poses~\cite{sun2023pose}.

\subsection{Rotation Averaging}\label{sec:litrev_rotationavg}
To recover the absolute camera orientations, some methods decouple in Bundle Adjustment (BA) rotation estimation from translation and scene reconstruction~\cite{Lee_2021_CVPR}; or improve on it by combining a global optimizer with fast view graph filtering and a local optimizer~\cite{9577752}. Other methods propose incremental approaches, prioritizing the averaging over camera pairs whose relative pose is supported by more triplets in the view graph, to increase robustness to outliers~\cite{lee2022hara}. 
Regarding GNN-based approaches, 
NeuRoRA~\cite{purkait2020neurora} combines a GNN that generates outlier scores and an estimated rotation error for each input relative pose (CleanNet), with a graph network that performs rotation averaging only on the inliers.
Subsequent works on multiple rotation averaging include GNNs that perform both initialization and optimization through a differentiable multi-source propagation module~\cite{yang2021end}, or iterative approaches like RAGO~\cite{li2022rago}, which learns to incrementally optimize relative pose estimates using only supervision on the relative poses. 
Drawing from the success of NeRFs in computer vision, other methods perform multiple motion averaging in the neural volume rendering framework ~\cite{jain2022robustifying}.
More recent works investigate the possibility of unsupervised models for rotation averaging, based on deep matrix factorization~\cite{tejus2023rotation} and directly propagating the uncertainty from the keypoint correspondences used in relative pose estimation, to more accurately model the underlying noise distributions~\cite{zhang2023revisiting}.

\subsection{Translation Averaging}\label{sec:litrev_translationavg}
Translation averaging uses as input pairwise relative translations and the (possibly) noisy absolute rotations from the previous rotation averaging step. 
Classical approaches~\cite{990963} predict absolute translation by optimizing the $l_2$ norm of the vector product between each input relative translation direction and the distance between the corresponding predicted absolute translations. Subsequent works improved results by relaxing the loss to $l_\infty$ ~\cite{moulon2016openmvg}, by introducing  a least unsquared deviations (LUD) form~\cite{zyesil2014RobustCL}, or by redefining the problems as the minimization of the projection of the absolute translations difference onto the relative translation direction~\cite{Goldstein2016ShapeFitAS}. These methods manage to frame the problem as the minimization of a convex cost function, but the scale ambiguity on each translation still requires extensive preprocessing strategies.
Other approaches consider instead angular residuals $\theta_{ij}$ between the predicted and observed translation directions, framing the task as a quasi-convex problem on $\tan \theta_{ij}$, solved through a sequence of Second Order Cone Programming (SOCP) problems~\cite{Sim2006RecoveringCM}. To reduce the impact of outliers, other works directly minimize a non-linear cost function of $sin \theta_{ij}$ via
Levenberg-Marquard~\cite{wilson_eccv2014_1dsfm} or iteratively reweighted Least Squares~\cite{zhuang2018baseline}.

\section{\APPROACHNAMEwithSpace for Objectness Refined Rotation Averaging}
We propose PRAGO, an end-to-end network for relative pose refinement and rotation averaging, aimed towards the estimation of absolute camera poses $[R^*, T^*]$. Our model takes as input  a set of images with matched objectness detections (Fig.~\ref{fig:main_fig}a) and, as proposed in PoserNet~\cite{taiana2022posernet},  we exploit information on matched detections to define covisibility, and employ the 5-point algorithm to generate relative pose estimation (Fig.~\ref{fig:main_fig}b). This results in a sparsely connected view graph, with noisy relative pose estimates which may contain outliers. 
First, we use objectness-based relative pose refinement (Fig.~\ref{fig:main_fig}c) to correct or refine outliers (Sec.~\ref{sec:posernetplus}). Then the refined poses are passed to a rotation averaging module (Fig.~\ref{fig:main_fig}d) to produce a set of absolute camera orientations (Sec~\ref{sec:rotationaveraging}). Both refined relative poses and absolute rotations are input to a translation averaging algorithm (Fig. \ref{fig:main_fig}e) to predict the absolute camera poses (Fig.~\ref{fig:main_fig}, Sec.~\ref{sec:translationavg}). To frame the problem, we first establish the preliminaries and notation in Sec~\ref{sec:methods_intro}, before defining the PRAGO components.


\subsection{Preliminary Problem Formulation}\label{sec:methods_intro}
Motion averaging is defined as the problem of recovering the absolute orientation $\hat{R}$ and translation $\hat{T}$ for a set of $N$ cameras, given a set or noisy observed relative orientations $R_{ij}$ and translations unit vectors $T_{ij}$.
For a pair of cameras $(i,j)$, given perfect relative transformations $\hat{R}_{ij}$ and $\hat{T}_{ij}$, their relations with absolute poses are defined as: 
\begin{align}
    \hat{R}_{ij} = \hat{R}_i \hat{R}^T_j, \; && \; \hat{R}_i \hat{T}_{ij} = \frac{\hat{T}_j - \hat{T}_i}{\|\hat{T}_j - \hat{T}_i\|_2}.
\end{align}
The observed relative poses are, however, affected by noise and can include outliers. 

A standard solution to motion averaging is to split the problem into two optimization steps, one for the rotations and one for the translations, computing the absolute poses more compatible with the observations. 
The absolute orientations $R^*$ are estimated by minimizing a robust cost function $\rho(\cdot)$ of the distance $d^R(\cdot)$ between $R_{ij}$ and the predicted $\tilde{R}_{ij} = \tilde{R}_{i} \tilde{R}^{T}_{j}$ as:
\begin{equation} \label{eq:ravg}
    R^*=\argmin_{R=\{ R_i \}} \sum_{(i,j) \in \mathcal{E}} \rho(d^R(R_{ij}, R_{i} R^T_{j})).
\end{equation}
The solution to this problem is, however, not unique. Applying the same rigid transformation to all absolute orientations $\tilde{R}$ does not change the relative orientations: $R_i R^T_j= R_i \bar{R} \bar{R}^T R^T_j = (R_i \bar{R}) (R_j \bar{R})^T$. To make the solution unambiguous, a fixed \emph{gauge transformation} is needed.

Given the predicted $R^*$, a cost function $\sigma$ and a vector distance $d^T(\cdot)$, the absolute translations are obtained as:  
\begin{equation} \label{eq:tavg}
    T^*=\argmin_{T=\{ T_i \}} \sum_{(i,j) \in \mathcal{E}} \sigma (d^T(T_{ij}, R^{* \; T}_{i} (T_j - T_i))).
\end{equation}
In Eq.~\ref{eq:ravg} and Eq.~\ref{eq:tavg}, the cost functions $\rho (\cdot)$ and $\sigma (\cdot)$ model the noise on the relative transformations; the same problems can be approached with a learning paradigm, training a network to implicitly model the noise. In this case, graph networks are especially well suited, as the problem can be naturally described as a graph $\mathcal{G} = (\mathcal{N}, \mathcal{E})$, where nodes $N_i\in \mathcal{N}$ represent the cameras, and edges $e_{ij} \in \mathcal{E}$ represent the observed relative poses between nodes $i$ and $j$. 


\subsection{Relative Pose Refinement}\label{sec:posernetplus}
To learn how to correct the relative pose outliers, we opt for a graph-based approach. Each camera $C^i$ is associated to a node, whose initial embedding is defined as $h=[f,H,W]$, where $f$ is the camera focal length and $H$ and $W$ are the height and width of the associated image. We then draw for each image pair $(i,j)$ a set of directed edges, for each pair of matched objectness detections $(l,m)$ with $l \in i, m\in j$. This is, in contrast, with~\cite{taiana2022posernet}, which draws a single edge between a pair of nodes, and handles the multiple detections on one node in an \textit{ad hoc} way. We define the edges embeddings by concatenating the relative pose between the cameras, $[R_{ij}, T_{ij}]$, and the bounding boxes $BB^i_{l}$ and $BB^j_{m}$ associated to the two detections. The former are represented as a quaternion for the rotation and as a unit vector for the translation. The latter are defined as $BB = [x,y,bb_w, bb_h]$, where $x$ and $y$ are the pixel location of the upper left bounding box corner and $bb_w$ and $bb_h$ respectively represent the bounding box's width and height in pixels.
This graph is passed through two GATv2~\cite{brody2022how} layers, modified to include the edge embeddings in the node's updates (drawing from Edge-GAT~\cite{101093bbab371}) and to apply attention. During this process, only the node embeddings are updated, while the edge embedding preserve information about the bounding boxes locations and the input relative poses. In contrast, \cite{taiana2022posernet} applies an \textit{ad hoc} GNN which updates both edge and node information.
To update the relative pose between camera pair $(i,j)$, we first use an MLP to merge together the concatenation of the learned embeddings of the nodes, $h_{ij}=\text{MLP}([h_i,h_j])$. Then, a second MLP takes as input the noisy relative pose and this joint embedding, to generate a refined pose as 
\begin{equation}
    \tilde{R}_{ij}, \tilde{T}_{ij} = \text{MLP}([R_{ij}, T_{ij}, h_{ij}]).
\end{equation}

Both the graph network and the two MLPs are trained to minimize the loss
\begin{equation}
    \mathcal{L}_{ref} = \mathcal{L}_{orient} + \mathcal{L}_{tr\_dir} + \gamma(\mathcal{L}_{q||} + \mathcal{L}_{tr||}),
\end{equation}
where $\mathcal{L}_{orient}$ and $\mathcal{L}_{tr\_dir}$ are angular losses, based respectively on the quaternion combination $\tilde{R}_{ij} \circ \hat{R}_{ij}$ and on $\cos^{-1} (\tilde{T}_{ij}\cdot \hat{T}_{ij}\|\hat{T}_{ij}\|^{-1})$; $\mathcal{L}_{q||}$ and $\mathcal{L}_{tr||}$ are respectively normalization losses on the quaternions and unit vectors; and $\gamma$ is a coefficient used to tune the contribution of the different components of the loss empirically set to $0.5$.

\subsection{Rotation Averaging}\label{sec:rotationaveraging}
Given the refined relative poses, we combine them to predict the best fitting absolute orientations $\tilde{R}$, drawing from the RAGO~\cite{li2022rago} architecture.
This module 
uses two Gated Recurrent Units (GRU) to iteratively update the global orientation of node $i$ and the relative orientations of its   nearest neighbors $\mathcal{N}_i$. First, the absolute poses $R_i$ are randomly initialized; then, $R_i$, $\tilde{R}_{ij}$ and $\mathcal{G}$ are passed to four MPNN to generate features $g$ for each node and edge, and initialization embeddings for the hidden states of the node and edges in the GRUs. At each optimization step $k=1, \dots K$, the cost of the current absolute and relative orientations for node $i$ are respectively $d_i^k$ and $d^k_{ij}$:
\begin{equation}
    d_i^k = \frac{1}{\mathcal{N}_i} \sum_{j \in \mathcal{N}_i} \| R^k_i - \tilde{R}_{ij}R^k_j \|.
\end{equation}
\begin{equation}\label{eq:outliercost}
    d^k_{ij} = \|R^k_{ij} - R^k_i R^{k\; T}_j \| + \| R^k_{ij}-\tilde{R}_{ij}\|.
\end{equation}
Here, $R^t_{ij}$ is an estimated relative pose, obtained by rectifying $R^k_i R^{k\; T}_j$ to align it to $\tilde{R}_{ij}$; the cost of Eq.~\ref{eq:outliercost} will then act as an outlier score, as it will tend to $0$ for inliers 
and will be large for outliers. 
Another MPNN extracts features $\mathcal{C}^k$ from the cost function, that are concatenated with the current orientation $R^k$ and the features into $I^k = [C^k, R^k,g]$. The current orientations can then be updated as: 
\begin{align}
    h^k = \text{GRU}(h^{k-1}, I^k) && \Delta R^k = \Phi(h^k),
\end{align}
where $h^k$ is the hidden state of the GRU nodes as time $k$, $\Phi$ is an MLP. $\Delta R^k$ is a correction to the orientation estimate, allowing to update the current predictions as
\begin{align}
    R^{k+1}_i = R^k_i \Delta R^k_i && R^{k+1}_{ij} = R^k_{ij} \Delta R^k_{ij}.
\end{align}
To speed up convergence, during training nodes and edges are optimized in alternated fashion; for a fixed number of node ($K_n$) and edge ($K_e$) iterations, the loss is given by:
\begin{equation}
\begin{split}
        \mathcal{L}_{avg} = \frac{1}{|\mathcal{E}|}\sum_{k=1}^{K_n} \sum_{(i,j) \in \mathcal{E}} \gamma^{K_n-k} \| R_i^T R_j^{k\; T} - \hat{R}_{ij} \|_1 + \\
    \frac{1}{|\mathcal{E}|}\sum_{k=1}^{K_e} \sum_{(i,j) \in \mathcal{E}} \gamma^{K_e-k} \| R_{ij}^T - \hat{R}_{ij}  \|_1.
\end{split}
\end{equation}
Note that this loss enforces supervision only on the ground-truth relative pose $\hat{R}_{ij}$, and not on the absolute poses. This avoids the gauge ambiguity issue. At convergence, the absolute rotation poses prediction is given by $R^* = \{R^K_i\}$.

The PRAGO components for pose refinement and rotation averaging modules are fully differentiable, and can be trained end-to-end by minimizing the final loss:
\begin{equation}
    \mathcal{L} = \alpha \cdot \mathcal{L}_{ref} + \beta \cdot \mathcal{L}_{avg},
    \label{eq:combined_losses}
\end{equation}
where $\alpha$ and $\beta$ are scaling coefficients. In practice, the residual on $\mathcal{L}_{ref}$ is approximately two orders of magnitude larger than $\mathcal{L}_{ref}$, and we set $\alpha = 1$, $\beta=0.01$.

\subsection{Translation Averaging}\label{sec:translationavg}
The relative poses from Pose Refinement and absolute orientations generated by Rotational Averaging are combined to predict absolute translations. For this, we apply BATA~\cite{zhuang2018baseline}, which frames translation averaging as an optimization problem of a bilinear objective function. BATA introduces a variable that performs normalization on the translations to base the loss on the angular error on the translation unit vectors, avoiding numerical instability due to scale ambiguity (Sec.~\ref{sec:litrev_translationavg}). This prevents translation vectors with a larger module from dominating the loss; in fact, at parity of noise on the orientation and on the module, the difference between ground truth and estimated translation vector will be proportional to the vector module. In contrast, a loss based on an angular distance will treat all relative translations similarly, regardless of the actual pairwise distance. The absolute translations of Eq.~\ref{eq:tavg} can be found by minimizing:
\begin{equation}\label{eq:bata}
     T^* = \argmin_{T, d} \sum_{(ij)\in \mathcal{E}} \mu (\| (T_j-T_i)d_{ij} - \tilde{T}_{ij}\|_2)
\end{equation}
where $\mu(.)$ is a robust M-estimator and $d$ is a non-negative variable, used to scale the module of $T_j-T_i$. This can be solved by including two constraints:
\begin{align}
  \sum_{i\in \mathcal{V}} T_i = 0  && \sum_{(ij)\in \mathcal{E}} \langle T_j-T_i, \tilde{T}_{ij} \rangle = 1,
\end{align}
which respectively impose a reference system and a scale.
Note that Eq.~\ref{eq:bata} essentially is a Least Squares objective embedded in an M-estimator to account for the presence of outliers in the input data, and can be efficiently resolved using an iteratively reweighted Least Squares scheme. As rotation averaging is typically considered more reliable than translation averaging, BATA uses the following residuals: 
\begin{equation}
    \epsilon = \sqrt{\| (T_j-T_i)d_{ij} - \tilde{T}_{ij}\|^2_2 + \| \tilde{R}_{ij} - R^*_{i} R^{* \; T}_{j}\|_2^2},
\end{equation}
to update the weights.
Combined with the absolute orientations, this results in full camera poses $[R^*, T^*]$, which we can compare against the ground truth poses $[\hat{R}, \hat{T}]$ to evaluate the full pipeline.

\section{Evaluation}

\begin{table*}[t]
    \scriptsize
    \centering
    \setlength\tabcolsep{3.8pt}
    \begin{tabularx}{\textwidth}{c X c c c c c c c c c c c c}
    \toprule
         &         & \multicolumn{6}{c}{Orientation}        & \multicolumn{6}{c}{Translation}               \\ \cmidrule(lr){3-8} \cmidrule(lr){9-14}
     & Pipeline & 3$^\circ$ & 5$^\circ$ & 10$^\circ$ & 30$^\circ$ & 45$^\circ$ & $\eta$ & 0.05 m& 0.1 m& 0.25 m& 0.5 m& 0.75 m& $\eta$ \\\midrule
     \multirow{3}{*}{\begin{tabular}{@{}c@{}}Non-fully \\ differentiable\end{tabular}}
     & PoserNet + EIG-SE3     &  
         28.15\%  & 
 \underline{60.26\%}  &  
         85.72\%  & 
         94.43\%  & 
         96.55\%  &       
  \underline{4.21$^\circ$}  &
         
          0.04\%   & 
          2.69\%   &   
         34.88\%   & 
 \textbf{76.21\%}   &  
 \textbf{94.19\%}   &       
\underline{0.33 m}   \\

& PoserNet + NeuRoRA + BATA     & 
         22.60\%  & 
         55.91\%  &  
         84.89\%  & 
         95.97\%  & 
         97.37\%  &       
          4.59$^\circ$  &
         
\underline{1.20\%}  & 
\underline{10.07\%}  &   
\underline{36.89\%}  & 
        66.68\%  &  
        85.77\%  &       
         0.35 m \\ 

 & PoserNet + RAGO + BATA      & 
 \underline{28.84\%}  & 
         59.68\%  &  
 \underline{87.28\%}  & 
 \underline{96.17\%}  & 
 \underline{97.53\%}  &       
          4.26$^\circ$  &
         
 \textbf{ 1.50\%}   & 
 \textbf{11.59\%}  &   
 \textbf{40.97\%}  & 
\underline{71.58\%}   &  
\underline{90.09\%}  &       
 \textbf{0.31 m}    \\  
\midrule
Differentiable
        & PRAGO (Ours) + BATA & 
 \textbf{37.88\%}  &  
 \textbf{65.59\%}  &  
 \textbf{89.29\%}  & 
 \textbf{96.32\%}  & 
 \textbf{97.56\%}  &       
 \textbf{ 3.77$^\circ$}  &
         
          0.34\%  & 
          4.47\%  &   
         33.23\%  & 
         67.84\%  &  
         89.41\%  &       
          0.36 m    \\ 
    \bottomrule
    \end{tabularx}
    \centering
    \caption{\label{tab:main_results_fully_diff_vs_non_fully_diff}
    The Table presents rotation and translation median errors $\eta$ and the percentage of test graphs that falls under a given error threshold for rotation ( 3$^\circ$, 5$^\circ$, 10$^\circ$, 30$^\circ$, 45$^\circ$) and translation (0.05 m, 0.1 m, 0.25 m, 0.5 m, 0.75 m). We evaluate three baseline approaches that use non-fully-differentiable pose refinement against \APPROACHNAME. Best and second best values for each column are highlighted in bold and underlined, respectively. }
\end{table*}


We first present results highlighting the effectiveness of \APPROACHNAMEwithSpace on real-world data, comparing our pipeline against baseline motion averaging approaches (Sec.~\ref{main_experiment}); then, we report on ablation tests that explore how to best combine the components of the loss of \APPROACHNAMEwithSpace and how our relative pose refinement compares against PoserNet~\cite{taiana2022posernet} (Sec.~\ref{ablations}). Finally, in Sec.~\ref{analyses}, we investigate how the relative pose accuracy changes after pose refinement, and how these changes affect 
the accuracy of absolute pose estimates.


\textbf{Dataset --} 
For the evaluation setting, we adopt the 7-scenes dataset~\cite{glocker2013real-time} following the methodology of ~\cite{taiana2022posernet}. In particular, we use the view graphs from the ``small graphs dataset'', which consists of $14000$ training graphs ($2000$ per scene) and $7000$ test graphs ($1000$ graphs per scene). 
Each graph's image contains objectness detections generated through a pre-trained Object Localization Network (OLN)~\cite{kim2022learning}, and matches defined by an oracle, using the intersection over union of the projection of one bounding box onto the other, for each possible pair of bounding boxes.
Each graph contains eight nodes, with a connectivity dictated by having at least three matched detections, and relative camera poses between the two images estimated using SuperGlue~\cite{sarlin2020superglue}, the five-point algorithm and RANSAC. This results in an average of $15.5$ edges per graph out of the possible $28$ edges ($55.36\%$ connectivity). We note that $3.8\%$ of the test graphs contain multiple components, which is not an issue for relative pose (as the focus of ~\cite{taiana2022posernet}) but a problem for absolute pose estimation. To overcome this issue, we present results on the $6731$ test graphs, which contain one single component, and re-evaluate the PoserNet baseline in ablation. For a full discussion of the object detections, matching and graph generation process see~\cite{taiana2022posernet}.


\begin{figure}
        \centering
    \begin{tabular}{c}
          \includegraphics[width=0.49\linewidth]{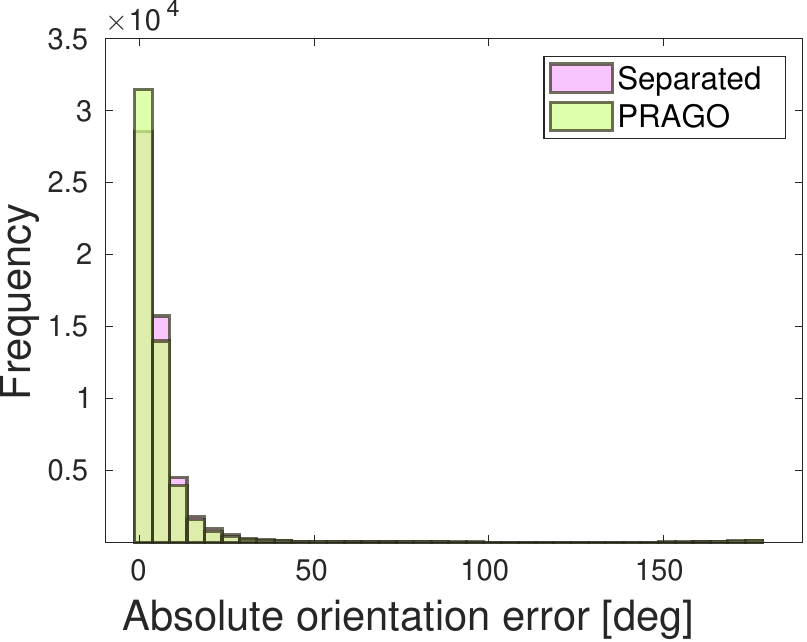}     
          \includegraphics[width=0.49\linewidth]{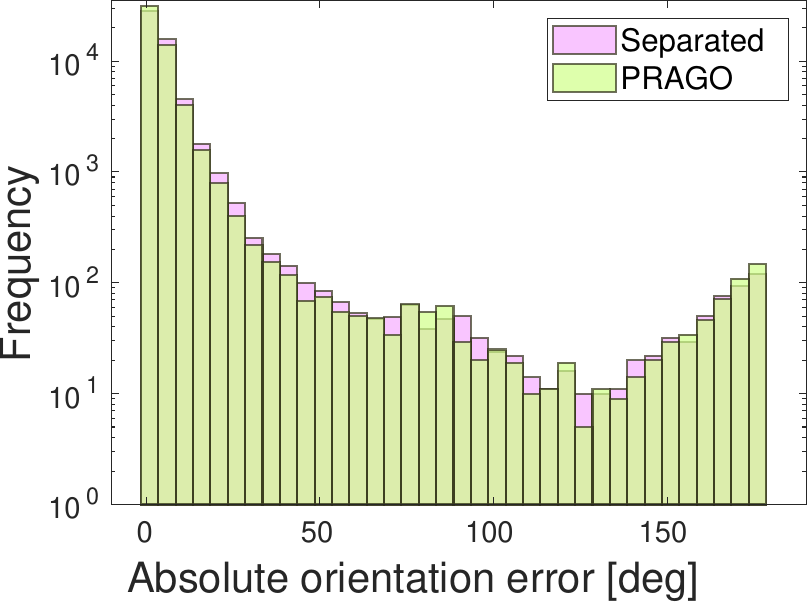}   
    \end{tabular}
    \caption{Histograms with natural (left) and logarithmic frequency axis (right) of absolute orientation errors produced by \APPROACHNAMEwithSpace (green) and by applying the pose refinement and the rotation averaging modules independently (pink). Histograms are overlaid with partial transparency.}\label
    {fig:separated_vs_joint_OptPoserNet_and_RAGO}
\end{figure}

\subsection{PRAGO Evaluation} \label{main_experiment}
We compare the performance achieved by the fully-differentiable \APPROACHNAMEwithSpace with that achieved by applying the pose refinement module and the rotation averaging module trained independently.
For reference, we also include baseline results obtained using the EIG-SE3~\cite{arrigoni2016spectral} motion averaging algorithm, or using the NeuRoRA~\cite{purkait2020neurora} rotation averaging method instead of RAGO. 
Results are reported in Tab.~\ref{tab:main_results_fully_diff_vs_non_fully_diff}, expressed in the form of the median error $\eta$ (the smaller the better) and as the percentage of testing graphs with and average error smaller than predefined thresholds (the higher the better). Such thresholds are $3$, $5$, $10$, $30$ and $45$ degrees for the angular metric, and  $0.05$, $0.1$, $0.25$, $0.5$ and $0.75$ for errors expressed in meters.
It can be seen that applying \APPROACHNAMEwithSpace achieves better orientation accuracy than the non-fully-differentiable baselines, both in terms of median error ($3.77^\circ$, against the second best of $4.21^\circ$ achieved by PoserNet followed by EIG-SE3) and in terms of percentage of graphs with errors below the specified thresholds. This improvement comes at the cost of a slightly lower accuracy for the estimated translations, with a median of $0.36 \:m$ compared with the $0.31 \:m$ achieved by running PoserNet, RAGO and BATA in a sequence.
The histogram of orientation errors obtained by applying \APPROACHNAMEwithSpace versus applying separately trained pose orientation and rotation averaging modules are shown in Fig.~\ref{fig:separated_vs_joint_OptPoserNet_and_RAGO}. It can be seen that \APPROACHNAMEwithSpace produces an overall better output distribution, with more errors concentrated bin corresponding to lowest error values.

\subsection{Ablations} \label{ablations}
To provide further insight on the design choices of the proposed pipeline, first we ablate across the hyper-parameters of \APPROACHNAMEwithSpace (Eq.~\ref{eq:combined_losses}); then, we compare our revised version of objectness pose refinement against PoserNet. \\
\textbf{Combining partial loss functions --} 
When training the fully differentiable \APPROACHNAME, we have the problem on how best to combine the losses as in Eq.~\ref{eq:combined_losses}.
We devised three strategies: driving the learning using exclusively the rotation averaging loss ($\alpha = 0, \beta=1$), driving it by naively adding the two loss components ($\alpha = 1, \beta=1$) and, finally, driving the learning by weighting the loss components so that they would have comparable values at the end of the training of the single components ($\alpha = 1, \beta=0.01$).
As can be seen in Tab.~\ref{tab:ablation_on_loss_aggregation}, relying exclusively on the rotation averaging loss leads to the overall best result in terms of rotation averaging, but this comes with a large penalty for translation averaging, with a median error almost twice as high as that achieved by the other loss combination schemes ($0.61 \:m$ versus $0.36 \:m$ and $0.32 \:m$). 

\begin{figure*}[t!]
    \centering
    \begin{tabular}{c}
          \includegraphics[height=0.24\linewidth]{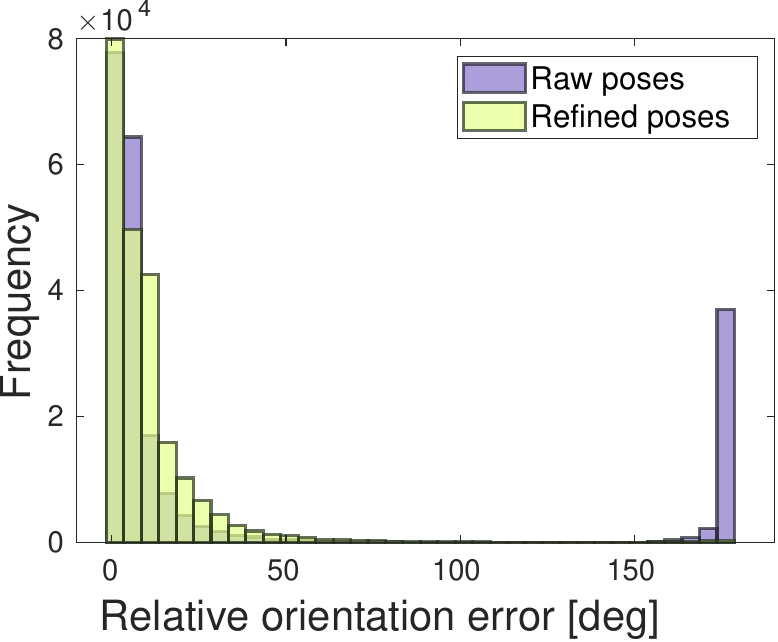} 
          
          \includegraphics[height=0.235\linewidth]{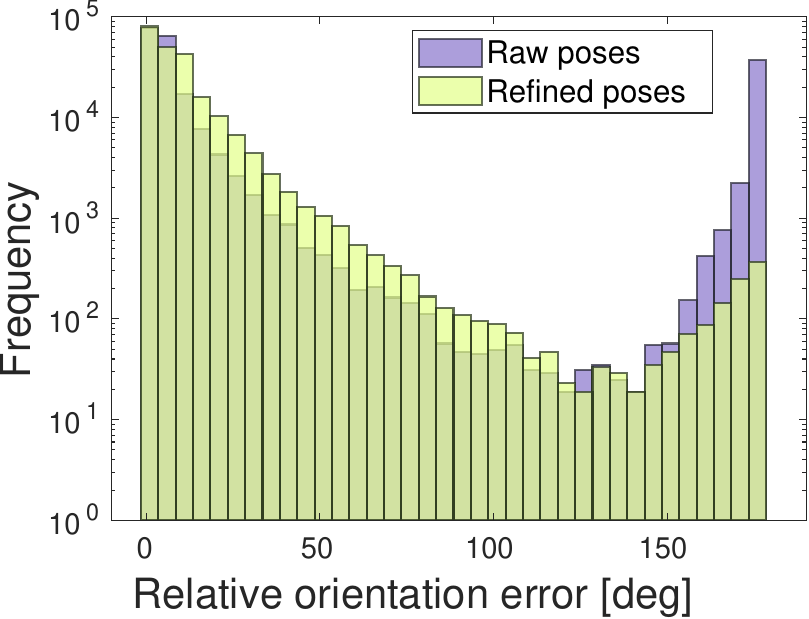}      
          \includegraphics[height=0.23\linewidth]{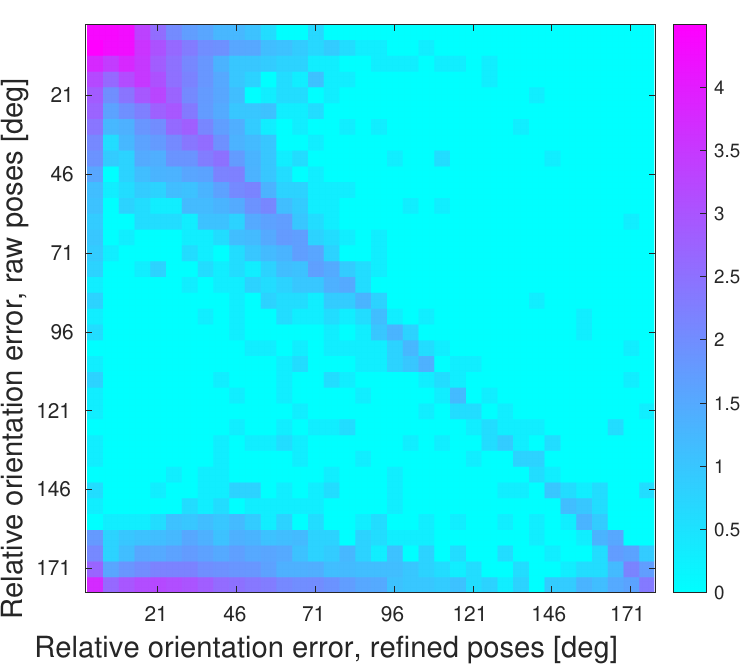}      
    \end{tabular}
    \caption{\emph{Effect of objectness-based pose refinement on the input relative poses.} Left and center, natural and logarithmic histograms of edge-wise orientation error, respectively, for the test set with raw initial poses (violet) and refined poses (green). Right, shows a matrix representing the refinement effect based on relative orientation error. The color represents $log_{10}$ of the number of graphs that belong to one cell of the matrix.}\label{fig:chirality_reduction}
\end{figure*}

\begin{table}[h]
    \scriptsize
    \centering
    \begin{tabularx}{\columnwidth}{c c c c }
    
    \toprule
    
    $\mathcal{L}_{ref}$  & $\mathcal{L}_{abs}$ & {Orientation error (deg)}        & {Translation error (meters)} \\
    \midrule

    0.00 & 1.00  &
         \textbf{3.75}  & 
         0.61   \\                 
    \midrule

    1.00 & 1.00  &
         3.77  & 
         0.36    \\                 
    \midrule

    1.00 & 0.01  &
         3.87  & 
         \textbf{0.32}    \\                 
    \bottomrule
    
    \end{tabularx}
    \centering
    \caption{\label{tab:ablation_on_loss_aggregation}
    Absolute orientation and translation errors achieved by using different weights for combining PRAGO's losses. Weights for each component are listed in the $\mathcal{L}_{ref}$  and $\mathcal{L}_{abs}$ columns.
    }
\end{table}

\begin{table}[h]
    \scriptsize
    \centering
    \begin{tabularx}{\columnwidth}{X c c }
    
    \toprule
    
    Pipeline              & {Orient. error (deg)}        & {Transl. error (meters)} \\
    \midrule

   {PoserNet + EIG-SE3}  &
         4.97  & 
         \textbf{0.32}   \\                 
    \midrule

    {Enhanced pose ref. + EIG-SE3}  &
         \textbf{4.21}  & 
         0.33    \\                 
    \bottomrule
    
    \end{tabularx}
    \centering
    \caption{\label{tab:old_PoserNet_vs_new_PoserNet}
    Absolute orientation and translation errors achieved by using either PoserNet or the proposed relative pose refinement architecture prior to applying EIG-SE3. }
\end{table}

\noindent\textbf{Enhanced objectness-based pose refinement --} We compare the performance obtained by EIG-SE3 when applied respectively on relative poses refined either by PoserNet or by the pose refinement architecture we propose.  As it can be seen in Tab.~\ref{tab:old_PoserNet_vs_new_PoserNet}, applying the proposed refinement architecture rather than PoserNet shows a drop in median orientation error from $4.97^\circ$ to $4.21^\circ$, while the corresponding impact on median translation error is negligible. 

\subsection{Analysis of the effect of pose refinement} \label{analyses}
In the initial formulation of objectness relative pose refinement in PoserNet~\cite{taiana2022posernet} showed the benefits of refining relative poses and, thus, leading to more accurate absolute poses. We explore this further to understand and explain a possible cause for this improvement firstly in its relative setting then in absolute when applying motion averaging.

\noindent\textbf{Refining relative poses with objectness --} 
We characterize the effect of the proposed enhanced pose refinement architecture has on estimates affected by different levels of noise. From the edge-wise orientation error histograms in Fig.~\ref{fig:chirality_reduction} it can be seen that pose refinement greatly reduces the number of estimates with orientation error larger than $160^\circ$, the ones which likely stem from an incorrect chirality choice.  
The transfer matrix on the right side of Fig.~\ref{fig:chirality_reduction} visualizes the effect pose refinement has on estimates affected by different levels of noise. Elements below and above the main diagonal correspond to estimates whose accuracy is increased or decreased, respectively. On the bottom-left corner of the matrix, it can be seen that many estimates with the wrong chirality are corrected to very accurate values (many of the input estimates that fall in the bin centered at $176^\circ$ are corrected to estimates that fall in the bin centered at $1^\circ$). This chirality-correction property is not perfect, though, as other such estimates are corrected to estimates with lower, but non-negligible levels of error (e.g., estimates that are moved from the bin centered on $176^\circ$ to that centered on $46^\circ$). 
Other effects are visible on the matrix, though perhaps not as clearly or not with a straightforward interpretation. On the top-left corner of the matrix, it can be seen that estimates with lower input errors can be affected both positively and negatively by the refinement. Finally, the high values present on the first column of the matrix indicate that many estimates are corrected to very low error values.   

This analysis highlights that objectness-based relative pose refinement has a strong positive impact on estimates affected by incorrect chirality, while also improving other estimates, suggesting that it could be especially important when coupled with motion averaging systems which do not implement an outlier rejection strategy.

\begin{figure*}[t]
        \centering
    \begin{tabular}{ccc}
          \includegraphics[width=42mm]{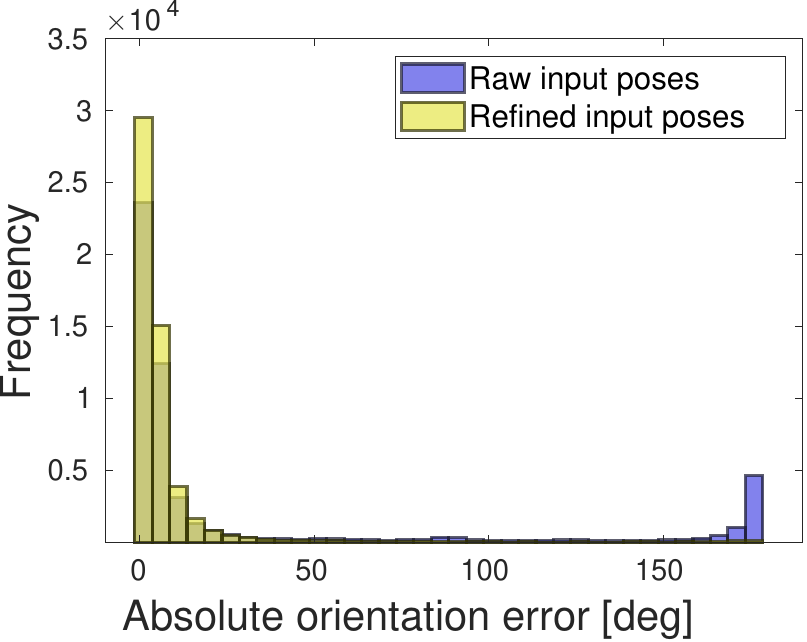}      &  
          \includegraphics[width=42mm]{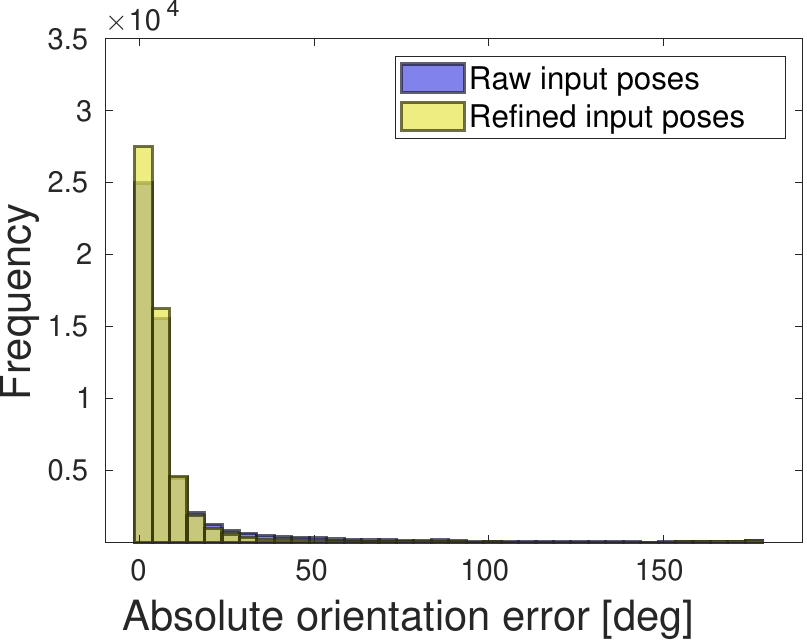}      & 
          \includegraphics[width=42mm]{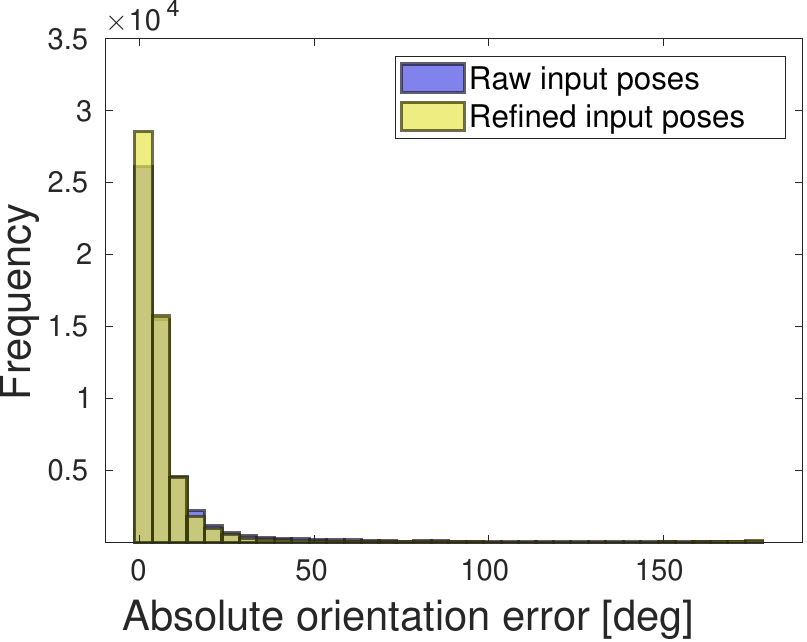}         \\

          \includegraphics[width=42mm]{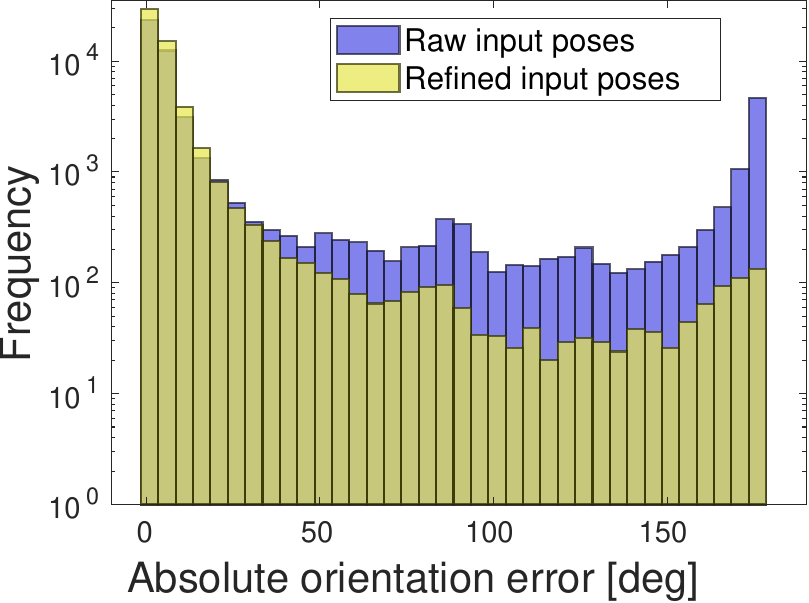}      &  
          \includegraphics[width=42mm]{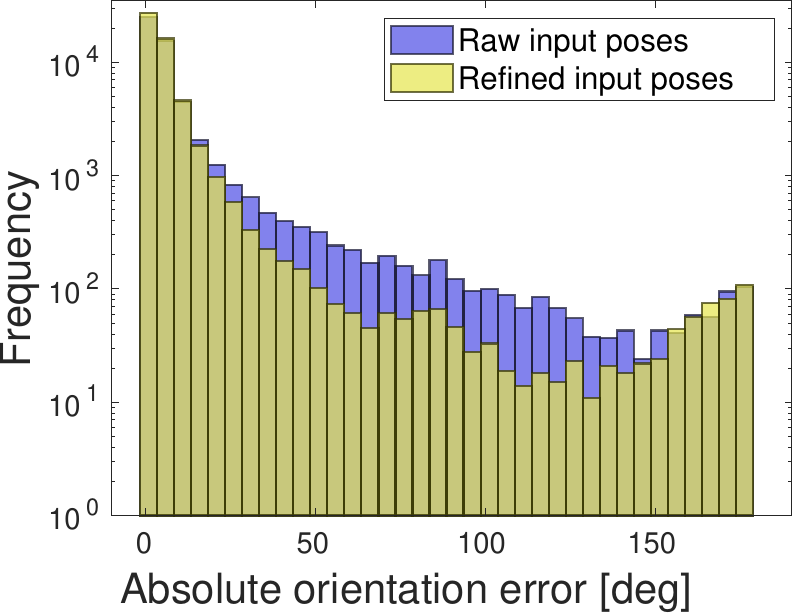} & 
          \includegraphics[width=42mm]{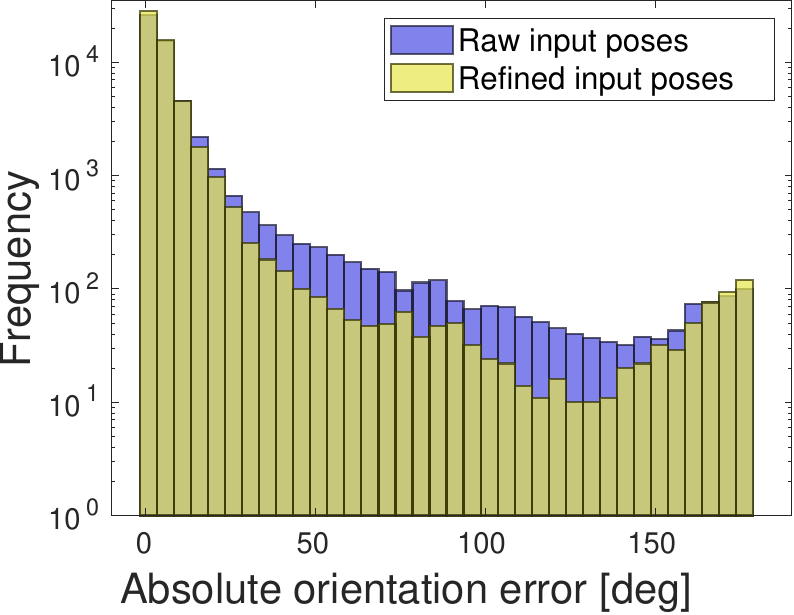}    \\
          
          \small (a) EIG-SE3 & (b) NeuRoRA &  (c) RAGO \\
    \end{tabular}
    \caption{Natural (top) and logarithmic (bottom) histograms of the absolute orientation error achieved by different rotation averaging methods (EIG-SE3, NeuRoRA and RAGO) applied either on raw relative poses (blue) or on refined relative poses (yellow). 
    }\label{fig:absolute_orientation_error_OptPoserNet_followed_by_x}
\end{figure*}

\begin{table}[t]
    \scriptsize
    \centering
    \begin{tabularx}{\columnwidth}{c X c c}
    \toprule
    \begin{tabular}{@{}c@{}}Motion Averaging \\ Pipeline\end{tabular}  & 
    \begin{tabular}{@{}c@{}}Type of \\ input poses\end{tabular}  & 
    \begin{tabular}{@{}c@{}}Orientation error\\ (deg)\end{tabular} &
    \begin{tabular}{@{}c@{}}Translation error\\ (meters)\end{tabular} \\\midrule
    \multirow{2}{*}{EIG-SE3} 
    & Raw   &    
         25.22  &
         
          0.61  \\
         \cmidrule(lr){2-4} 
                  
     & Refined     &  
  \textbf{4.21}  &
         
          0.33   \\
         
    \midrule
    \multirow{2}{*}{NeuRoRA, BATA} & Raw      & 
          5.84  &
         
          0.61  \\  \cmidrule(lr){2-4}

         & Refined    & 
          4.59  &
         
         0.35  \\ 
         
    \midrule
    \multirow{2}{*}{RAGO, BATA} & Raw      & 
          5.09  &
         
          0.60    \\  \cmidrule(lr){2-4}
         
         & Refined & 
          4.26  &
         
 \textbf{0.31}    \\  
         
    \bottomrule
    \APPROACHNAMEwithSpace (Fully diff.) & Raw      &        
          3.77  &
         
          0.36    \\  
    
    \bottomrule
    \end{tabularx}
    \centering
    \caption{\label{tab:effect_of_pn_on_motion_averaging_error}
    Impact of relative pose refinement on absolute poses. Orientation and translation error are shown for different motion averaging pipelines fed with either raw or refined relative poses. The performance of \APPROACHNAMEwithSpace is shown for reference.}
\end{table}

\noindent\textbf{Impact of relative pose refinement on Motion Averaging --} 
To understand the cumulative effect of applying objectness relative pose refinement, we compare the performance achieved by pairing the proposed pose-refinement architecture with three different Motion Averaging systems: EIG-SE3, NeuRoRA followed by BATA, and RAGO followed by BATA. 
It is important to notice that EIG-SE3 is not trainable, so it is applied without any customization to the problem. For NeuRoRA, we fine-tuned its two pre-trained networks (CleanNet and FineNet) on the training data from 7-Scenes as provided in~\cite{taiana2022posernet}. We trained RAGO from scratch on the same training data and applied it on the test data directly. The authors of RAGO mention that they preprocess real data with NeuRoRA's CleanNet, but for our experiments, we opted to apply it directly on raw relative poses or to the refined ones. 

We report the orientation and translation errors on raw and refined relative poses in Tab.~\ref{tab:effect_of_pn_on_motion_averaging_error}, and show the corresponding histograms of orientation errors in Fig.~\ref{fig:absolute_orientation_error_OptPoserNet_followed_by_x}. The results show that applying objectness-based pose refinement is beneficial for all motion averaging algorithms, with the median translation error dropping from roughly $0.6 \:m$ to roughly $0.3 \:m$ for all of them. In terms of absolute orientation errors, EIG-SE3 is the algorithm which benefits the most from the relative pose refinement. When applied on raw relative transformations, EIG-SE3 produces the estimates with the highest median rotational error ($25.22^\circ$), highlighting a sensitivity to highly noisy input data. As can be seen in Fig.~\ref{fig:absolute_orientation_error_OptPoserNet_followed_by_x}a, EIG-SE3 is not able to solve all the chirality errors present in the raw input data (a considerable number of its output estimates have errors close to $180^\circ$). 
When applied on refined relative poses (where most very high errors have been reduced), instead, EIG-SE3 produces the lowest median orientation error of all methods ($4.21^\circ$), narrowly beating RAGO.

\subsection{Implementation details}
As the input relative poses, we used the five-point poses from~\cite{taiana2022posernet}. We implemented the architecture for objectness-based pose refinement using the PyTorch framework~\cite{pytorch} and PyTorch Geometric~\cite{FeyLenssen2019pyg}, and trained it using the Adam optimizer with learning rate $0.0001$ and batch size $1$. We used a schedule to reduce the learning rate by a factor of $0.316$ with a patience of $3$ epochs. We used the author implementation for each of the other methods
, EIG-SE3\cite{arrigoni2016spectral},  RAGO~\cite{li2022rago}, NeuRoRA~\cite{purkait2020neurora} and BATA~\cite{zhuang2018baseline}. For training or fine-turning those methods, we used the training set defined in~\cite{taiana2022posernet}, and the hyperparameters present in their code or defined in their papers. For the training \APPROACHNAME, we started from versions of the pose refinement and rotation averaging models trained in isolation, and fine-tuned the joint system using the same hyperparameters described earlier. 

\section{Conclusion}
We have presented PRAGO, a novel differentiable camera pose estimation approach exploiting objectness detections. Our method takes as input a set of unordered images with their corresponding detections to estimate initial pair-wise relative poses. The pairwise rotations are then refined 
 with a novel approach that can work efficiently with rotational averaging iterative optimization blocks. By differentiating through the method, we have shown that the PRAGO method can achieve superior performance in contrast to concatenating its pre-trained components.
In addition, through analysis, we have shown that a key cause of the performance increase is improving the pair-wise orientation errors, especially in the case of chirality, during pose refinement, a step largely unconsidered by large-scale techniques with abundant images. PRAGO has demonstrated the benefits of learning in an end-to-end fashion, as opposed to focusing on the optimization of individual tasks alone.




{
    \small
    \bibliographystyle{ieeenat_fullname}
    \bibliography{main}
}

\end{document}